\documentclass[letterpaper, 10 pt, conference]{ieeeconf}  % Comment this line out if you need a4paper
\usepackage{xcolor}
\usepackage{graphicx}
\usepackage{amsfonts}
\usepackage{hyperref}
\usepackage{amsmath}

\IEEEoverridecommandlockouts                              % This command is only needed if 
\title{\LARGE \bf
Contextual Data Integration for Bike-sharing Demand Prediction with Graph Neural Networks in Degraded Weather Conditions}
\author{Romain Rochas$^{1}$, Angelo Furno$^{1}$ and Nour-Eddin El Faouzi$^{1}$% <-this % stops a space
\thanks{This work was not supported by any organization}% <-this % stops a space
\thanks{$^{1}$Romain Rochas, Angelo Furno and Nour-Eddin El Faouzi are with LICIT-ECO7 UMR T9401, University Gustave Eiffel, University of Lyon, ENTPE, Lyon, France.
        {\tt\small romain.rochas@entpe.fr}, {\tt\small angelo.furno@univ-eiffel.fr}, {\tt\small nour-eddin.elfaouzi@univ-eiffel.fr}}%
}

\begin{document}

\maketitle
\thispagestyle{empty}
\pagestyle{empty}

%%%%%%%%%%%%%%%%%%%%%%%%%%%%%%%%%%%%%%%%%%%%%%%%%%%%%%%%%%%%%%%%%%%%%%%%%%%%%%%%
\begin{abstract}
Demand for bike sharing is impacted by various factors, such as weather conditions, events, and the availability of other transportation modes. This impact remains elusive due to the complex interdependence of these factors or location-related user behavior variations. It is also not clear which factor is additional information which are not already contained in the historical demand. Intermodal dependencies between bike-sharing and other modes are also underexplored, and the value of this information has not been studied in degraded situations. The proposed study analyzes the impact of adding contextual data, such as weather, time embedding, and road traffic flow, to predict bike-sharing Origin-Destination (OD) flows in atypical weather situations
Our study highlights a mild relationship between prediction quality of bike-sharing demand and road traffic flow, while the introduced time embedding allows outperforming state-of-the-art results, particularly in the case of degraded weather conditions. Including weather data as an additional input further improves our model with respect to the basic ST-ED-RMGC prediction model by reducing of more than 20\% the prediction error in degraded weather condition.
\end{abstract}

%%%%%%%%%%%%%%%%%%%%%%%%%%%%%%%%%%%%%%%%%%%%%%%%%%%%%%%%%%%%%%%%%%%%%%%%%%%%%%%%
\section{INTRODUCTION}
\label{sec:INTRODUCTION}
%Papier sur la qualité de prédiction de vélo dans le cas de scénario météo. La part scientifique de mon papier, c'est l'ajout de donnés contextuelles météo, multi-modale, et embedding de temps, et  pour voir si cela améliore la qualité de prédiction des vélos 
In recent years, there has been a significant surge in the adoption of soft mobility options, driven by factors such as health considerations, escalating oil prices, and personal ecological concerns. This growing trend towards soft mobility, which includes biking and other forms of non-motorized transportation, has prompted the need for accurate demand prediction on the network. Such prediction improvements hold substantial benefits for both users and operators of Bike-Sharing Systems (BSS), particularly in addressing challenges like bike rebalancing \cite{Hua M1}.
In practice, the bike-sharing systems serve as connection on the multi-modal transport network \cite{C. Zhang1}, allowing for reduced travel times, cost-effectiveness, and a smaller spatial and environmental footprint compared to traditional modes of transportation \cite{R. Buehler1}.
While demand forecasting has historically relied on statistical and machine learning techniques, the advent of deep learning has ushered in a new era of forecasting algorithms. Among these, GNNs \cite{Bruna1} appeared in 2014, but it has only been applied to the transportation domain from 2017 \cite{Yu. B1} and has emerged as the state-of-the-art models for capturing spatial dependencies in transportation prediction. Various models exist for bike-sharing demand prediction, ranging from station level forecast \cite{Chai. D1}, to cluster-based prediction \cite{Chen. L1}, and many other variations deeply detailed in surveys such as \cite{Jiang. W1}.
However, only a limited number of studies have focused on Origin-Destination (OD) prediction \cite{Li. Y2}. OD prediction poses unique challenges, as the final destinations are usually unknown, and OD matrices generated are sparse \cite{Zhang et1, Whang et1, KE2021102858}. Moreover, forecasting demand in the case of atypical phenomena, such as non-recurrent events, remains a challenge \cite{An. J1}).
Bike-sharing demands exhibit temporal and calendar dependencies tied to factors such as weekdays, public holidays, workdays, and school holidays \cite{c1}. Furthermore, they are influenced by weather conditions such as wind, humidity, temperature, and particularly rainfall \cite{c1},\cite{c2},\cite{c4}. While this introduction does not aim to comprehensively address all the factors impacting bike-sharing demand, in-depth studies provide further insights  \cite{Eren. E1}.
%\inter{Les études ? ? ? mettent en avant des liens multimodaux entre demande en vélo et demande en voiture sur le réseau. }
%\inter{Egalement, une forte pluie impact la demande jusqu'à 3h après sa fin, \cite{c3}, ce qui justifie l'utilisation d'un embedding de temps sur le nombre d'heure depuis la fin de la pluie.}
We have chosen to work on an OD prediction algorithm, the ST-ED-RMGC \cite{KE2021102858} as it is one of the few addressing the issue of forecasting by OD, it is a recent state-of-the art model, and it integrates proven prediction modules (Multi-Graph-Convolution, Encoder-Decoder and Residual Module) for traffic prediction. In contrast to the original paper, we have chosen to work with bike-sharing data, incorporating contextual information such as time, weather conditions, and multi-modal data, including road car flow. Specifically, we evaluate the inclusion of contextual data for predicting Weather-related scenarios. We have already worked on this algorithm to make an empirical analysis of the forecasting accuracy of the algorithm when predicting bike-sharing demand under the constraint of weather scenarios. This paper goes further by integrating contextual data within the training phase. For further information, the reader can refer to our last paper \cite{ICASP14}. 
The main contributions of this study are the following:
\begin{itemize}
    \item We propose the integration of an embedding module in the ST-ED-RMGC to capture the calendar-based dependencies of bike-sharing demand.
    \item We propose the integration of contextual weather data to account for the meteorological sensitivity of bike-sharing demand.
    \item We propose the integration of road car flow data by zone to capture multi-modal dependencies under rainy scenarios.
    \item We evaluate the addition of these contextual data not only on the global prediction performances of the neural network, but also on the prediction quality during atypical weather scenarios 
\end{itemize}
The rest of the paper is organised as follows. In Sec.~\ref{sec:METHODOLOGY}, we present the problem being addressed, introduce the ST-ED-RMGC model and describe the proposed improvements by adding embedding and contextual data. In Sec.~\ref{sec:DATASET}, we present the available dataset and the spatial aggregation applied to road loop detectors and bike sharing stations. Sec.~\ref{sec:EXPERIMENTS}, details the results obtained from the various scenarios tested. Sec.~\ref{sec:CONCLUSION} concludes the study while outlining future research directions.
\section{METHODOLOGY}
This section details the problem and the methodology proposed initially in \cite{KE2021102858}, as well as our proposed enhancements to the original model that leverage contextual data in several forms to enhance the prediction accuracy in degraded weather conditions.

\label{sec:METHODOLOGY}
\subsection*{Original Problem Statement}
\label{subsec:Problem statement}
%\textcolor{red}{DEBUT reprise sur papier ICASP14 - Modfié }
Let $\mathcal{Z}$ represent the set of areas corresponding to a partitioning of the analyzed territory, and 
let $OD_{\mathcal{Z}} = \{ (z_i, z_j) | z_i, z_j \in \mathcal{Z},  i \neq j\}$ denote the set of all pairs of zones. We define $Y^{t}_{(z_i,z_j)}\in \mathbb{R}$ as the bike-sharing demand between  $(z_i,z_j)$ at $t$ time-step. Specifically, $Y^{t}_{(z_i,z_j)}$ represents the outflow demand for destination $z_j$ that left from $z_i$ during time slot $t$.
 Let $^{B}X^{t}_{(z_i,z_j)} = \left[Y^{t-7d}_{(z_{i},z_{j})},Y^{t-d}_{(z_{i},z_{j})},Y^{t-2}_{(z_{i},z_{j})},Y^{t-1}_{(z_{i},z_{j})}  \right] $ be the associated sequence of previously observed bike-sharing demands which allows to predict  $Y^{t}_{(z_i,z_j)}$, where $t-2$ and $t-1$ represent respectively the period $t$ minus $1$ and $2$ time-steps, and $d$ represent a time-step of 1 day.
 Let $G_{u}$ be a complete graph associated to a relationship $u$ on $OD_{\mathcal{Z}}$, and 
 let $A_{u}$ be its adjacency matrix which takes into account a spatial dependency. Multiple $u$ relationships can be jointly considered, and $\ddot{A}$ denotes the concatenation of all these adjacency matrices. All the considered adjacency matrices are detailed in \ref{subsec: Adjacency Matrices}. The temporal dependencies are instead taken into account via the historical demand sequence data and the inherent periodicity of such data. 
 Let $Y^{t} = \left[ Y^{t}_{(z_{i_{1}},z_{j_{1}})},...,Y^{t}_{(z_{i_{N}},z_{j_{N}})} \right] \in \mathbb{R}^{N \times 1}$, where $N$ is the number of ODs, and  $^{B}X^{t} = \left[ ^{B}X^{t}_{(z_{i_{1}},z_{j_{1}})},...,^{B}X^{t}_{(z_{i_{N}},z_{j_{N}})} \right] \in \mathbb{R}^{N \times 4}$. The problem can be described as follows: 
\begin{equation}
    Y^{t}= F(^{B}X^{t},\ddot{A})  
\end{equation}
where $F$ is  the prediction function.
%\textcolor{red}{FIN reprise sur papier ICASP14}
\subsection{ST-ED-RMGC} 
\begin{figure}[h!]
      \centering
      \framebox{\parbox{3in}{
      \centering
      \includegraphics[scale=0.3]{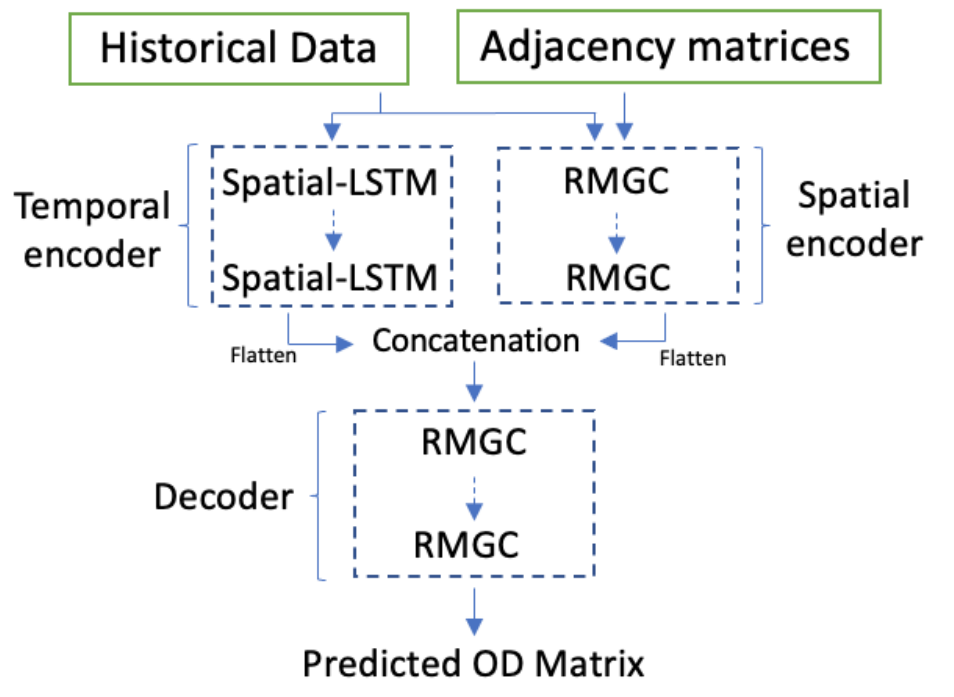}}}
      \caption{Framework of the ST-ED-RMGC model, adapted from  \cite{KE2021102858}}
      \label{fig:ST_ED_RMGC_framework}
    \vspace{-10pt}
    \end{figure}

    Figure~\ref{fig:ST_ED_RMGC_framework} displays the overall architecture of the model. The model is an encoder-decoder based model, where the encoder is composed by a temporal encoder which takes into account the temporal dependencies of all OD pairs (and not just one at a time), and a spatial encoder with several Residual-Multi-Graph-Convolutional network (RMGC) which takes several adjacency matrix and the graph OD demand as inputs. The RMGC combines a residual module with a multi-graph convolution, to capture the spatial correlation between OD pairs. The residual module is introduced to tackle the issue of gradient explosion/gradient vanishing in complex deep networks. The multi-graph convolution applies graph convolution on stacked weighted adjacency matrices. 
    \par
\subsection{Adjacency Matrices}
\label{subsec: Adjacency Matrices}
The relationships introduced in Sec.~\ref{subsec:Problem statement} can refer to a spatial relationship or a semantic relationship. As proposed in \cite{KE2021102858}, a spatial relationship between two OD pairs can be expressed via two relationships: an origin-based, and a destination-based, denoting a relationship on the origins (respectively destinations) of the two OD pairs. The neighbourhood ($A_{n}^{O}, A_{n}^{D}$) and centroid distances  ($A_{d}^{O},A_{d}^{D}$) are used as spatial relationships. We defined semantic relationships in the same way, where a semantic relationship between two areas is not dependent on their geographical location. Thus, the functionality between two areas is translated into an origin-based and destination-based functionality adjacency matrix  ($A_{f}^{O}, A_{f}^{D}$). The functionality of an area is represented by a vector containing the socio-economic information of an area, such as the presence of a railway station, or the density of housing. A relationship on the correlation between the historical demands of two OD pairs is also defined ($A_{corr}$). In this paper, we then define seven different adjacency matrices, which are all concatenated in $\ddot{A}$ to capture the dependencies between zones. 
\begin{equation} \label{concatenated adjacency matrix}
\ddot{A} = \left[ A_{n}^{O}, A_{n}^{D}, A_{d}^{O},A_{d}^{D},A_{f}^{O} A_{f}^{D},A_{corr} \right] 
\end{equation}
\subsection{Road sensor data integration}
Differently from the original ST-ED-RMGC approach, which only uses bike-sharing historical data as the main input to the forecasting model, we also decided to integrate, as an additional input, information related to the (historical) car traffic flow, as observed in the proximity of the origin of the OD bike-sharing flow we are trying to predict. The main idea is that this information could allow capturing the inter-dependence between car and bike usages, which could be helpful to anticipate, \emph{e.g.,} modal changes such as people switching from bikes to cars in the presence of degraded weather conditions or other kinds of perturbations. From an architectural perspective, the data from road sensors are concatenated to $^{B}X^{t}_{(z_i,z_j)}$ $\forall i,j,k$.  Let $I^{t}_{i}$ be the total car traffic flow recorded in zone \textit{i} during time slot $t$.
We define a new feature vector: 
\begin{equation} \label{Intensity feature vector}
^{I}X^{t}_{(z_{i},z_{j})} =  \left[ ^{B}X^{t}_{(z_{i},z_{j})},I^{t-T_{I1}}_{i},...,I^{t-T_{I2}}_{i} \right] \forall i,j
\end{equation}
where $(T_{I1}-T_{I2}+1)$ is the depth of historical car flow observations considered, with $T_{I1} > T_{I2}$
\subsection{Weather data integration}
\label{sec:weather_data}
In the same way, we decided to integrate information related to the historical weather, as well as weather forecast. Knowing that bike-share use is impacted before, during or after rain, the interest here is in capturing patterns in changes in bike-share use related to degraded weather.
Let $W^{t}$ a sequence of features from the weather dataset, which will be concatenated to $^{B}X^{t}_{(z_{i},z_{j})} \forall i,j$. $W^{t}$ can be composed of historical data such as $\left[v^{t-T_{W1}},...,v^{t-T_{W2}} \right]$, where \textit{v} can be set as \textit{hr}, corresponding to the hourly rainfall ($mm.h^{-1}$), or \textit{hd}, corresponding to the hourly rainfall duration $(min$ per hour$)$, depending on the considered model variation. $(T_{W1}-T_{W2}+1)$ is the number of weather historical time-steps provided as an input. Note that $T_{W1} > T_{W2}$ and $T_{W2}$ can be negative or zero, which corresponds to the use of a weather prediction for the next  $-T_{W2}$ time-steps. $W^{t}$ can also contain weather forecast such as $dcr^{t}$ $(mm.h^{-1}$ per day$)$, \emph{i.e.,} the daily cumulative rainfall recorded the day associated to the date $t$. 
We thus define the new feature vector: 
\begin{equation} \label{Weather feature vector}
^{W}X^{t}_{(z_{i},z_{j})} =  \left[ ^{B}X^{t}_{(z_{i},z_{j})},W^{t} \right] \forall i,j.
\end{equation}
We can finally define a new feature vector including both contextual weather and sensor information as follows: 
\begin{equation} \label{complete feature vector}
^{W,I}X^{t}_{(z_{i},z_{j})} =  \left[ ^{B}X^{t}_{(z_{i},z_{j})},W^{t},I^{t-T_{I1}}_{i},...,I^{t-T_{I2}}_{i} \right] \forall i,j
\end{equation}
\subsection{Time Encoding}
We have integrated a time-encoding component to take into account temporal contextual information. Demand for bike sharing, and more generally, any transport mode, is calendar-dependent. It is normally linked to working hours, and therefore directly related to the day of the week or the time of the day. Demand in university zones is for instance impacted by school vacations. Similarly, in business districts, employees with children can take vacations accordingly. For these reasons, we decided to integrate different calendar-related features as contextual data:  weekday, whether we are trying to predict a business day, a school holiday, a day of departure for school holidays or a return day from school holidays. Each of these \textit{r} features have been encoded as a binary vector $T_{i}$ whose dimension is equal to the number of classes in the feature \textit{i}, except for the feature "hour". As for the feature "hour", we have labelled the 24 possible hours in 18 classes. From 1 to 17 for the first 17 hours from 6 a.m., and 0 for the others. It avoids adding unnecessary information, as the bike ODs during the night hours are almost zero. As for the others features, they have been labelled from 0 to the number of possible classes (\emph{e.g.,} 0,1,..6 to label the 7 days of the week). Then, each labelled features has been encoded in a binary vector of the dimension of the number of classes in the labelled feature. The final output of this module is a list of vectors: $T_{1},...,T_{r}$.

%all the labelled feature have been encoded in a binary vector of the dimension of the max of the labels +1. The final output of this module is a list of vectors: $T_{1},...,T_{r}$.
%As for the feature "hour", we have labelled the 24 possible hours in a vector of dimension 17. From 1 to 17 for the first 17 hours from 6am, and 0 for the others. It avoids adding unnecessary information, as the bike ODs during the night hours are almost zero. 
%Then, all the labels were encoded in a binary vector of the dimension of the max of the labels +1 \fixme{What do you call "max of the labels"?}. The final output of this module is a list of vectors: $T_{1},...,T_{r}$.
\subsection{Time Embedding}
\begin{figure}[thpb]
      \centering
      \framebox{\parbox{3in}{
      \includegraphics[scale=0.35]{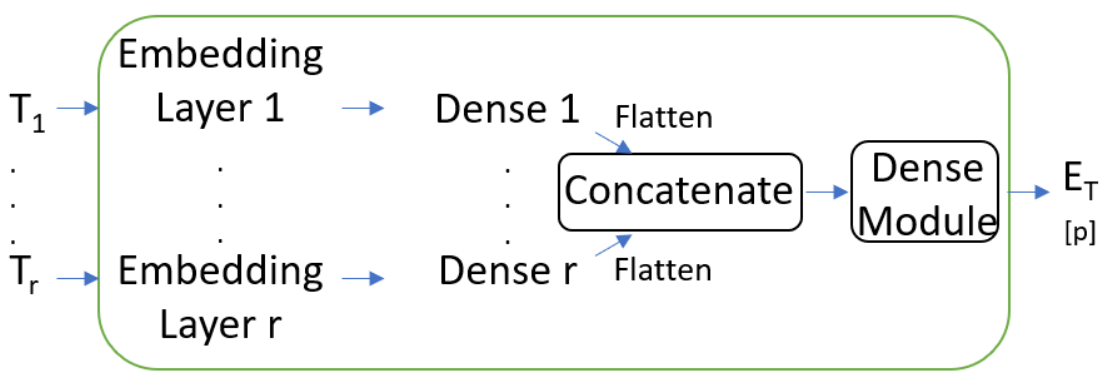}}}
      \caption{Embedding module. The Dense Module is composed of 3 dense layers in series.}
    \vspace{-10pt}
      \label{fig:Embedding_module}
   \end{figure}

\begin{figure}[thpb]
      \centering
      \framebox{\parbox{3in}{
      \includegraphics[scale=0.26]{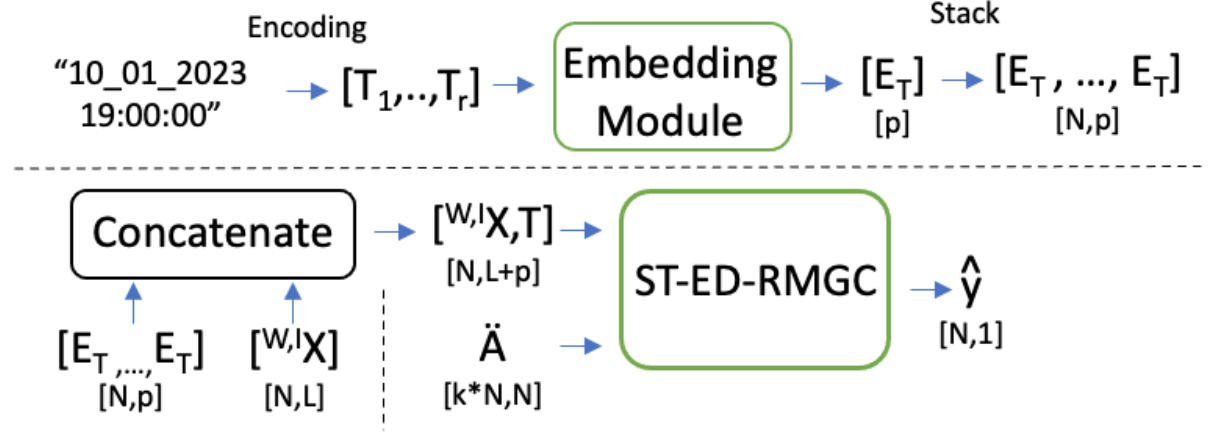}}}
      \caption{Integration of the embedding module into the existing architecture.}
    \vspace{-10pt}
      \label{fig:Integration_of_the_embedding_module}
   \end{figure}

We have chosen to implement an Embedding module (see Fig~\ref{fig:Embedding_module}) to take into account temporal contextual information and the dependencies between them. The embedding module contains \textit{r} embedding layers in parallel, each of them corresponding to a unique $T_{i}, i \in \left[1,r\right]$. Each of the outputs of the embedding layers passes through a dense layer, before being concatenated and finally passing into a dense module (composed of 3 dense layers). Dense layers capture the dependencies between each of the time-encoded vectors. The output of the dense module is $E_{T}$, a vector of dimension \textit{p}. This vector represents the temporal information of the date \textit{t} that we are trying to predict. As the time embedding is the same for each ODs (\emph{i.e.,} the time embedding does not capture spatial dependencies), $E_{T}$ is then stacked as many times as the number of selected ODs, which gives a 2D matrix $\left[ E_{T},...,E_{T} \right]$ of size [\textit{N,p}]. Finally, this matrix is concatenated (Fig~\ref{fig:Integration_of_the_embedding_module})
to the feature vector $X$ to obtain a new feature vector of dimension [\textit{N,L+p}].

\section{Datasets}
\label{sec:DATASET}
\subsection{Bike-sharing Data}
We work with bike-sharing data provided by JC Decaux. We have actual drop-in and drop-out data per station. Working by OD between stations does not make sense because the dynamic OD matrix will be sparse and would not represent user mobility. Stations have physical limitations: a finite set of spatially localised stations with limited vehicle capacity. Many users, therefore, use the bikes available at stations within a certain radius of their departure area to reach a certain radius of their arrival destination. It was thus necessary to aggregate the bike share data not only temporally, but also spatially. The stations were first aggregated by IRIS zones (Fig~\ref{fig:aggregation}), which are zones developed by the French Institute of Statistics. This segmentation divides the conurbation of Lyon into small geographical areas, each grouping approximately $2\,000$ inhabitants\footnote{\url{https://www.insee.fr/fr/metadonnees/definition/c1523}}. The IRIS areas are thus used to group bike-sharing stations according to socio-economic criteria.  To reduce the sparsity of the data, we then decided to further aggregate pairs of IRIS zones according to an iterative procedure described below, based on three criteria while maintaining the spatial homogeneity of the division. After this aggregation, we end up with a very sparse OD matrix (because all zones are connected). We filtered the ODs by setting a proportion $p_{\text{bike}}$ of total demand between 7 a.m. and 9 p.m. and taking the minimum number of ODs needed to reach this proportion. 
\subsection{Spatial Aggregation}
\label{sec:spatial_aggreg}
\begin{figure}[thpb]
  \centering
  \framebox{\parbox{2.8in}{  %3in
  \includegraphics[scale=0.184]{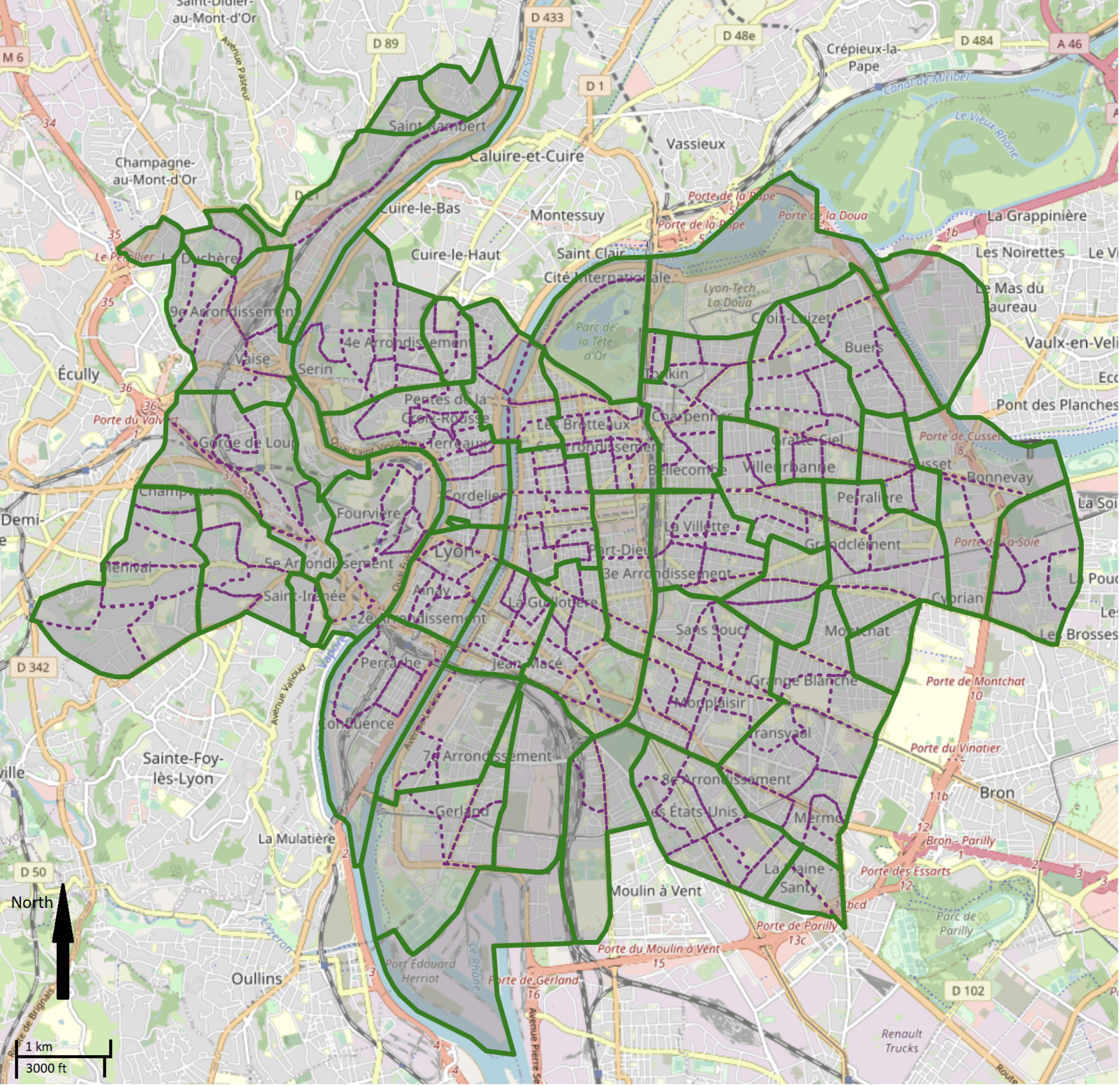}}} %0.195
  \caption{IRIS zones without any aggregation (dashed line) and the aggregation into 50 zones (solid line). }
  \label{fig:aggregation}
\end{figure}

The adopted aggregation (Fig~\ref{fig:aggregation}) procedure builds upon the following three criteria: the proximity between two IRIS zones, the sum of their area and the common perimeter between two adjacent areas. 
Specifically, at each iteration, we look for the two IRIS zones $i^*$ and $j^*$ minimizing the following objective: 
\begin{equation} \label{spatial aggregation}
(i^*,j^*) = argmin_{(i,j) \in Z_{\textnormal{adj}}}\Big(\Big\{ \frac{1}{P_{i,j}}(s_{i} + s_{j}) \Big\} \Big)
\end{equation}
where $\mathcal{Z}_{\textnormal{adj}}$ is the set of adjacent pairs of zones in the current spatial aggregation, $s_{i}$ the surface of the zone $i$, and $P_{i,j}$ is the common perimeter between $i$ and $j$. Zones are aggregated up to a fixed number of zones.
\vspace{-10pt}
\subsection{Car Flow Data}
Data from loop detectors provide information on the flow and loop occupancy rate of cars, recorded on the loop. Our initial dataset provided counts aggregated every 6 minutes. 
The data were first aggregated by hour and by loop, keeping a sample if the loop recorded more than 4 periods of 6 minutes in an hour. A period is kept if it does not represent erroneous data, \emph{i.e.,} an abnormal loop occupancy rate ($>50\%$), or a NaN value flow. Otherwise, we remove the data, as they are not statistically correct. Each of the aggregated areas contains several loops. Data were then aggregated by zone. At the end of the preprocessing, only about 10 periods of 1 hour could not be determined. For such hours we simply adopted a linear interpolation. 
\subsection{Weather Data}
For the integration of the weather context as described in Sec.~\ref{sec:weather_data}, we leveraged data provided by Meteo France\footnote{\url{https://meteofrance.com}} and collected hourly from 2 weather stations of the city of Lyon, reporting weather conditions (\emph{e.g.,} hourly rainfall intensity and duration, temperature, \emph{etc.}) from January 1st, 2019, to December 31st, 2020. The two stations have been aggregated by taking the average of the values on both. Weather data are therefore independent of the zone, but dependent on the hourly time slot $t$.  
\section{EXPERIMENTS}
\label{sec:EXPERIMENTS}
The train dataset used in our experiments includes hours from January 8th, 2019 to November 9th, 2019, while the test dataset covers the 7 a.m. and 9 p.m. period for all days between December 1th, 2019, and March 10th, 2020. The 220 IRIS zones from Lyon and Villeurbanne have been aggregated into 50 zones (Fig~\ref{fig:aggregation}), which correspond to 2500 OD pairs. We decided to perform the prediction exclusively on the minimum number of ODs necessary to represent $p_{\text{bike}}= 60\%$ of the bike-sharing demand from 7 a.m. to 9 p.m. This meant that we only worked with 130 of the 2500 ODs of our territory. The algorithm was trained on all the hours of the day (contrary to the test scenarios limited between 7 a.m. and 9 p.m.). 

\subsection{Weather-related Test Scenarios}
%\fixed{}\fixme{you should introduce these scenarios: what are they useful for? Why do we consider them? How do they relate to the train/testing datasets?}
In addition to the global test dataset defined above and used to evaluate the overall performance of each model, we identified \emph{weather-related scenarios} to assess the accuracy of bike-sharing demand prediction in degraded weather conditions. Specifically, we used such test scenarios to comparatively evaluate the proposed models, which include contextual data, with respect to the baseline ST-ED-RMGC model that does not include them. The weather-related scenarios are built as subsets of the global test dataset, based on the daily cumulative rainfall levels ($dcr$) as well as the hourly amount of rain ($hr$). Indeed, the user's choice of bike sharing as travel mode can be strongly influenced by an ongoing rain episode, an expected rainfall, as well as an already concluded rainfall episode \cite{c3}. We thus include in the weather-related scenarios hours (between 7 a.m. and 9 p.m.) with rain ($hr > 0$), without rain ($hr = 0$), and with daily cumulative rainfall falling within specific positive ranges, as detailed in Tab~\ref{table: Weather-related scenarios}. 
%\fixed{en fait avec cette phrase pour être cohérent j'aurais du définire des scénario où il ne pleut pas mais qu'il va y avoir de la pluie dans le future/des scénario où il ne pleut pas mais il y a eu de la pluie dans le passé }\fixme{I am not sure to understand the previous sentence, and particularly the link that you want to make between the influence of weather forecast/concluded rain episode and the ranges in your table on the cumulative rain...}.
\begin{table}[!h]
\caption{Weather-related scenarios}%Rainfall scenarios (\fixme{Number of dates is actually "Number of hours"?}.}
\label{table: Weather-related scenarios}
\vspace{-13pt}
\begin{center}
\resizebox{3.3in}{!}{
\begin{tabular}{|c c|c|c||c c|c|c|}
\hline
n$^{\circ}$ & scenario & nb of  & \% of 0  & n$^{\circ}$ & scenario & nb of  & \% of 0 \\
&  &  hours &  values & &  &  hours &  values \\
\hline
\hline
0 & test dataset between  & 1500 & 12 & 3 & $dcr = 0$ & 960 & 11\\
 & 7 a.m. and 9 p.m.  &  & & & & & \\
\hline
1 & $hr = 0$ & 1344 & 12 & 4 & $dcr \in \left]0,1 \right]$  & 120 & 11\\
\hline
2 & $hr > 0$ & 156 & 20 & 5 & $dcr \in \left]1,3 \right]$ & 240 & 13\\
\hline
\end{tabular}
}
\end{center}
\vspace{-20pt}
\end{table}

\subsection{Model Settings}
The architecture parameters, such as the number of neurons and the number of layers, are those proposed in the original ST-ED-RMGC paper \cite{KE2021102858}.
%\fixed{}\fixme{what do you mean by "architecture parameters"?}
The default values of other hyperparameters in our experiments are set up as follows. We set the length of a time-step to 1 hour. The optimizer used in the model is Adam with a learning rate of $5e^{-5}$ and a decay of $1e^{-6}$. We set the dropout to $0.7$, the batch size to $16$, and the number of epochs to $80$ for all models. The loss function used to train the model is based on the Mean-Squared Error (MSE). These hyperparameters were chosen by validation with respect to the reference model, corresponding to the basic ST-ED-RMGC corresponding to the first line in Tab~\ref{table: Performance comparison}. All other models were trained with these same hyperparameters, ensuring that each of them was fully trained with no overfitting. In Tab~\ref{table: Performance comparison}, the column \emph{Model Type} denotes the type of embedding or contextual data which have been included (or not) within a specific model. \textit{X} denotes the usual feature vector without any additional information, \textit{T} denotes the one with time embedding, \textit{W} the one with weather data, \textit{I} the one with road car flow data. \textit{WIT} indicates a combination of \textit{W}, \textit{I}, and \textit{T}. The column \textit{Model Number} represents the number associated to the model variation being considered. \emph{e.g.,} the 5 models with type \textit{I} will be called \textit{I1}, \textit{I2}, \textit{I3}, \textit{I4}, and \textit{I5}. The column \textit{Included Context Elements} corresponds to the detailed description of the contextual elements considered within the model variation. 

\subsection{Metrics}
The metrics used to evaluate the models are the MSE and the Mean Absolute Percentage Error (MAPE). %Since the error is demand-dependent, and demand varies during the day, 
%The MAPE metric was considered to assess the prediction accuracy relatively to the demand magnitude. This makes it possible to compare prediction quality rather than an average of squared deviations. Some real OD demand is sometimes zero, in which case it is not considered in the calculation of the MAPE of a scenario. 
The OD matrix contains a large number of demands close to 0, that is why the MAPE is high $(> 0.5)$, since many predictions are close in absolute terms but the error is relatively high (e.g. real demand equal to 1 and prediction equal to 3).
We highlight here that the metrics depend on the demand. 
%\fixed{ fixed by remove a sentence ...} \fixme{You have to pay attention with this sentence as the MAPE is normalised with respect to the demand (as you have just said in the previous sentence on MAPE). So there seems to be a contradiction with respect to the MAPE metric here.}
It is therefore not possible to directly compare the performance of two different scenarios with each other, because the related datasets are different in terms of bike-sharing demand. It is nonetheless possible to compare models between them on the same scenario, as detailed below.

\subsection{Results}
Tab.~\ref{table: Performance comparison} reports the prediction results (\textit{MSE} and \textit{MAPE}) for scenario $n^{\circ}0$ from Tab.~\ref{table: Weather-related scenarios}
%, \emph{i.e.,} on the global test dataset without weather constraints, between 7am and 9 p.m. 
Tables~\ref{table: Model with rainfall data}:\ref{table: Time embedding model comparison} report instead the performance on the weather-related scenarios, with specific focus on the model variations including weather-related (Tab.~\ref{table: Model with rainfall data}), flow-related (Tab.~\ref{table: Model with intensity data}) and time embedding (Tab.~\ref{table: Time embedding model comparison}) features, respectively.
%\fixed{In the following subsection, am I wrong ?}\fixme{Where the results related to scenario 0 are described? You should comment these results anyway!}
\begin{table}[!h]
\caption{Overall performance comparison}
\label{table: Performance comparison}
\vspace{-20pt}
\begin{center}
\resizebox{3.4in}{!}{
\begin{tabular}{|c c||c||c||c|}
\hline
Model Type & Model Number & Included Context Elements & MSE & MAPE \\
\hline
\hline
X &  &  & 8.70 &  0.570 \\
\hline
\hline 
T &  & $E_{T}$ & 8.22 & 0.532 \\
\hline 
\hline 
W & 1 & $hr^{t-1}$ & 8.63 & 0.560\\  % W21
\hline 
 & 2 & $hr^{t-2},hr^{t-1}$ & 8.90  & 0.582 \\ %W15
\hline
 & 3 & $hr^{t-3},hr^{t-2},hr^{t-1}$ & 8.62 & 0.549  \\ %W20
\hline
 & 4 & $hr^{t}$ & 8.67  & 0.583 \\  %W24
\hline
 & 5 & $hr^{t-1},hr^{t}$ & 8.66  & 0.553 \\  %W25
 \hline
% & 16 & $hr^{t},hr^{t+1}$ & 8.58  & 0.560 \\
%\hline
% & 12 & $hr^{t-2},hr^{t-1},hr^{t},hr^{t+1}$ & 8.68  & 0.559 \\
%\hline
 & 6& $hr^{t-1},hd^{t-1}$  & 8.86  & 0.575 \\  %W9
\hline
 & 7& $hr^{t},hd^{t}$ & 8.56 & 0.55 \\  %W8
%\hline
% & 10 & $dcr^{t}$ & 8.61  & 0.551 \\
\hline
\hline 
I & 1 & $I^{t-1}$ & 8.78  & 0.588 \\ %I26
%& 6 & $Id^{t-1}, Ir^{t-1}$ & 8.64  & 0.537 \\
\hline 
& 2& $I^{t-2},I^{t-1}$ & 8.70 & 0.572 \\ %I13
\hline
& 3 & $I^{t}$  & 8.68  & 0.562 \\  %I22
\hline
& 4 & $I^{t-1},I^{t}$  & 8.47  & 0.555 \\  %I23
\hline
& 5 & $I^{t-2},I^{t-1},I^{t}$  & 8.44  & 0.562 \\  %I14
\hline
%& 27 & $Id^{t-1},Ir^{t-1}$  & 9.18 & 0.588  \\
%\hline
%& 28 & $Id^{t},Ir^{t}$  & 8.79 & 0.585 \\
%\hline
%& 29 & $Id^{t-1},Id^{t},Ir^{t-1},Ir^{t}$  & 8.85 & 0.581 \\
%\hline
\hline 
% WI & 17 & $hr^{t-2},hr^{t-1},hr^{t},hr^{t+1},I^{t-2},I^{t-1},I^{t} $ & 8.45  & 0.560 \\
%\hline
%& 18 & $hr^{t-1},hr^{t},hr^{t+1},I^{t-1},I^{t}$  & 8.46  & 0.551 \\
%\hline
\textbf{WIT} &  & \textbf{\textit{hr$^{\textbf{t-1}}$, hr$^{\textbf{t}}$, hr$^{\textbf{t+1}}$, I$^{\textbf{t-1}}$, I$^{\textbf{t}}$, E$_{\textbf{T}}$}} & \textbf{8.21} & \textbf{0.528} \\
\hline
%& \textbf{31} & \textbf{\textit{hr$^{\textbf{t-1}}$, hr$^{\textbf{t}}$, hr$^{\textbf{t+1}}$, Id$^{\textbf{t-1}}$, Id$^{\textbf{t}}$,Ir$^{\textbf{t-1}}$, Ir$^{\textbf{t}}$, E$_{\textbf{T}}$}} & 8.23 & \textbf{0.514} \\
%\hline
\end{tabular}
}
\end{center}
\vspace{-20pt}
\end{table}
\subsubsection{Model with weather features}

\newcommand\mcc[1]{\multicolumn{1}{c}{#1}}
\begin{table}[!b]
\caption{Model with weather-related data}
\vspace{-10pt}
\label{table: Model with rainfall data}
\begin{center}
\resizebox{3.3in}{!}{
\begin{tabular}{c c|c c || c c|c c }
\hline
scenario & model & mse & mape & scenario & model & mse & mape\\
\hline
\hline
$hr= 0$ & X & 8.71 & 0.551  & $dcr \in \left]0,1 \right]$ & X & 8.57 & 0.578\\
\hline
& W1 & 8.69 & 0.547 & & W1 & 8.44 & 0.565\\
\hline
& W2 & 8.94 & 0.568 & & W2 & 8.86 & 0.59\\
\hline
& W3 & 8.7 & 0.537 & & W3 & 8.66 & 0.553\\
\hline
& W4 & 8.87 & 0.58 & & W4 & 8.47 & 0.597\\
\hline
& W5 & 8.8 & 0.548 & & W5 & 8.68 & 0.562\\
\hline
& W6 & 8.92 & 0.563 & & W6 & 8.5 & 0.577\\
\hline
& W7 & 8.8 & 0.55 & & W7 & 8.48 & 0.562\\
\hline
& WIT & \textbf{8.33} & \textbf{0.521} & & WIT & \textbf{7.9} & \textbf{0.538}\\
\hline
\hline
$hr> 0$ & X & 8.68 & 0.747 & $dcr \in \left]1,3 \right]$ & X & 8.7 & 0.607 \\
\hline
& W1 & 8.1 & 0.688 & & W1 & 8.79 & 0.597\\
\hline
& W2 & 8.59 & 0.716 & & W2 & 9.21 & 0.619\\
\hline
& W3 & 7.99 & 0.663 & & W3 & 8.85 & 0.579\\
\hline
& W4 & 6.97 & 0.611 & & W4 & 8.43 & 0.605\\
\hline
& W5 & 7.48 & 0.606 & & W5 & 8.47 & 0.571\\
\hline
& W6 & 8.35 & 0.689 & & W6 & 9 & 0.605\\
\hline
& W7 & \textbf{6.88} & \textbf{0.546} & & W7 & 8.29 & 0.562\\
\hline
& WIT & 7.15 & 0.597 & & WIT & \textbf{8.14} & \textbf{0.541} \\
\hline
\hline
$dcr =  0$ & X & 8.86 & 0.541 & \mcc{} & \mcc{} & \mcc{} &\\
\cline{1-4}
& W1 & 8.84 & 0.539 & \mcc{} & \mcc{} & &\\
\cline{1-4}
& W2 & 9.06 & 0.56 & \mcc{} & \mcc{} & &\\
\cline{1-4}
& W3 & 8.8 & 0.531 & \mcc{} & \mcc{} & &\\
\cline{1-4}
& W4 & 9.03 & 0.569 & \mcc{} & \mcc{} & &\\
\cline{1-4}
& W5 & 8.95 & 0.538 & \mcc{} & \mcc{} & &\\
\cline{1-4}
& W6 & 9.07 & 0.556 & \mcc{} & \mcc{} & &\\
\cline{1-4}
& W7 & 8.96 & 0.541 & \mcc{} & \mcc{} & &\\
\cline{1-4}
& WIT & \textbf{8.52} & \textbf{0.518} & \mcc{} & \mcc{} & &\\
\cline{1-4}
\end{tabular}
}
\end{center}
\end{table}
Both in the case of scenarios without rain ($hr = 0, dcr = 0$, Tab.~\ref{table: Model with rainfall data}), and in the scenario $n^{\circ}$ 0 (Tab.~\ref{table: Performance comparison}), models with weather features show no significant change in the prediction quality compared to the baseline simpler model (X). However, in scenarios with rainfall ($hr > 0$), all models with weather information outperform the simple model. The model (W7) achieves the lowest MSE and MAPE in the scenario ($hr > 0$, Tab.~\ref{table: Model with rainfall data}), with approximately 20\% and 27\% improvement, respectively, compared to the (X) model. Notably, for this scenario, the (W7) model outperforms all other models, surpassing the second-best model (WIT) by 4\% in MSE and 9\% in MAPE. This means that the duration of the rain is directly linked to the bike-sharing demand. Compared with the other models, the (W7) model knows how long the rain will last over the next hour, and therefore the amount of time it will not rain in the next hour. There is strong reason to believe that as soon as the rain stops, a certain proportion of users will quickly resume their regular BSS use. For example, if employees who use bike-sharing are in the office, they will wait a few minutes for the rain to stop before moving.
\subsubsection{Model with flow features}
Tab~\ref{table: Model with intensity data} and Tab~\ref{table: Performance comparison} occasionally exhibit a slight improvement in the prediction quality with respect to the baseline (X) model. The improvement ranges from 0\% to 6\% in MSE and 0\% to 3\% in MAPE, depending on the models and scenarios (I1, I2, I3, I4, I5). However, considering the marginal improvement achieved by integrating these features in the specific analysed scenarios, it is not possible to draw conclusive evidence regarding the usefulness of their utilization. 
\begin{table}[!h]
\caption{Model with flow data}
\vspace{-10pt}
\label{table: Model with intensity data}
\begin{center}
\resizebox{3.3in}{!}{
\begin{tabular}{c c|c c || c c|c c }
\hline
scenario & model & mse & mape & scenario & model & mse & mape\\
\hline
\hline
$hr= 0$ & X & 8.71 & 0.551 & $hr> 1$& X & 7.22 & 0.978\\
\hline
% & I6 & 8.67 & \textbf{0.52}&  & I6 & \textbf{6.12} & \textbf{0.882}\\
& I1 & 8.78 & 0.568 & & I1 & 7.91 & 1.05\\
\hline
& I2 & 8.7 & 0.553 & & I2 & 7.19 & 0.991\\
\hline
& I3 & 8.71 & 0.543 & & I3 & \textbf{6.78} & \textbf{0.974} \\
\hline
& I4 & 8.51 & \textbf{0.535} & & I4 & 7.03 & 0.999\\
\hline
& I5 & \textbf{8.45} & 0.542 & & I5 & 7.31 & 0.985\\
\hline
\hline
$hr> 0$  & X & 8.68 & 0.747 & \mcc{} & \mcc{} & &\\
\cline{1-4}
%& I6 & 8.38 & \textbf{0.694}& \mcc{} & \mcc{} & &\\
& I1 & 8.87 & 0.78 & \mcc{} & \mcc{} & &\\
\cline{1-4}
& I2 & 8.68 & 0.745 & \mcc{} & \mcc{} & &\\
\cline{1-4}
& I3 & 8.4 & 0.741 & \mcc{} & \mcc{} & &\\
\cline{1-4}
& I4 & \textbf{8.12} & \textbf{0.737} & \mcc{} & \mcc{} & &\\
\cline{1-4}
& I5 & 8.44 & 0.745 & \mcc{} & \mcc{} & &\\
\cline{1-4}
\end{tabular}
}
\end{center}
\vspace{-15pt}
\end{table} 

\subsubsection{Model with time embedding} 
The application of time embedding results in an overall improvement in the results compared to the simple model (X), with lower MSE and MAPE in the order of 5.5\% and 6.6\%, respectively. Upon examining Tab~\ref{table: Time embedding model comparison}, it can be noted that predictions are superior regardless of the scenario. A scenario without rain demonstrates improved performance, with a reduced MSE and MAPE of 4.8\% and 6.2\%, respectively. The improvement is even more pronounced in scenarios with rainfall. For example, dates with rainfall presence ($hr > 0$) exhibit a significantly smaller MSE and MAPE, reduced by 12.4\% and 10.8\% respectively.
\begin{table}[!h]
\caption{Time embedding model comparison}
\vspace{-14pt}
\label{table: Time embedding model comparison}
\begin{center}
\resizebox{3.3in}{!}{
\begin{tabular}{c c|c c || c c|c c }
\hline
scenario & model & mse & mape & scenario & model & mse & mape\\
\hline
	$hr= 0$ & X & 8.71 & 0.551 & $dcr =  0$  & X & 8.86 & 0.541 \\
\hline
	 & T & \textbf{8.29} & \textbf{0.517} & & T & \textbf{8.49} & \textbf{0.515} \\
\hline \hline 
$hr> 0$   & X & 8.68 & 0.747 & \mcc{} & \mcc{} & & \\
\cline{1-4}
	& T & \textbf{7.6} & \textbf{0.666}& \mcc{} & \mcc{} & & \\
\cline{1-4} 
\end{tabular}
}
\end{center}
\vspace{-20pt}
\end{table}

\section{CONCLUSION AND FUTURE WORK}
\label{sec:CONCLUSION}
The results of this study emphasize the relevance of contextual data in OD bike sharing demand forecast, especially during degraded weather-related situations.
In future work, we plan to study scenarios without rain ($hr = 0$) among the days where a certain amount of rain has occurred ($dcr$ scenarios within a positive interval). This would enable studying the extent to which users are impacted by the risk or occurrence of rain and assess whether weather data can better anticipate user choices, even when it ultimately does not rain. The inter-modal relationships between (car) traffic flow in a specific area and BSS demand have not been extensively explored. A finer zoning approach might lead to more interesting conclusions. Finally, further research in this direction includes focusing on other inter-modal travel demand relationships, such as public transportation and BSS demand. 

%A conclusion section is not required. Although a conclusion may review the main points of the paper, do not replicate the abstract as the conclusion. A conclusion might elaborate on the importance of the work or suggest applications and extensions. 

\addtolength{\textheight}{-12cm}   % This command serves to balance the column lengths

%%%%%%%%%%%%%%%%%%%%%%%%%%%%%%%%%%%%%%%%%%%%%%%%%%%%%%%%%%%%%%%%%%%%%%%%%%%%%%%%

%%%%%%%%%%%%%%%%%%%%%%%%%%%%%%%%%%%%%%%%%%%%%%%%%%%%%%%%%%%%%%%%%%%%%%%%%%%%%%%%

%%%%%%%%%%%%%%%%%%%%%%%%%%%%%%%%%%%%%%%%%%%%%%%%%%%%%%%%%%%%%%%%%%%%%%%%%%%%%%%%
%\section*{APPENDIX}

\section*{ACKNOWLEDGMENT}

This research is supported by the French ANR research projects PROMENADE (grant number ANR-18-CE22-0008) and MOBITIC (grant number ANR-19-CE22-0010).

%%%%%%%%%%%%%%%%%%%%%%%%%%%%%%%%%%%%%%%%%%%%%%%%%%%%%%%%%%%%%%%%%%%%%%%%%%%%%%%%

\end{document}